\documentclass[11pt]{article}

\usepackage[final]{acl}
\usepackage{ragged2e}
\usepackage{times}
\usepackage{latexsym}
\usepackage{amsfonts}
\usepackage{amsmath}
\usepackage{makecell}
\usepackage[most]{tcolorbox}
\usepackage{orcidlink}
\usepackage{fontawesome5}

\usepackage[T1]{fontenc}

\usepackage{booktabs}
\usepackage{tabularx}  
\usepackage[dvipsnames,table]{xcolor}   

\usepackage[utf8]{inputenc}

\usepackage{microtype}

\usepackage{inconsolata}

\usepackage{graphicx}
\usepackage{subcaption}

\title{Rhetorical Questions in LLM Representations: A Linear Probing Study}

\author{
 Louie Hong Yao\textsuperscript{1},~ 
 Vishesh Anand\textsuperscript{2},~ 
 Yuan Zhuang\textsuperscript{3},~ 
 Tianyu Jiang\textsuperscript{2}
\\
 \textsuperscript{1}Independent Researcher\qquad
 \textsuperscript{2}University of Cincinnati\qquad
 \textsuperscript{3}Amazon
\\
 {\fontsize{10.2}{12}\selectfont
\texttt{lhyao731@gmail.com, anandvh@mail.uc.edu, zyone@amazon.com, tianyu.jiang@uc.edu}
}
}

\begin{document}
\maketitle

\begin{abstract}
Rhetorical questions are asked not to seek information but to persuade or signal stance. How large language models internally represent them remains unclear. We analyze rhetorical questions in LLM representations using linear probes on two social-media datasets with different discourse contexts, and find that rhetorical signals emerge early and are most stably captured by last-token representations.  Rhetorical questions are linearly separable from information-seeking questions within datasets, and remain detectable under cross-dataset transfer, reaching AUROC around 0.7-0.8. \textit{However, we demonstrate that transferability does not simply imply a shared representation.} Probes trained on different datasets produce different rankings when applied to the same target corpus, with overlap among the top-ranked instances often below 0.2. Qualitative analysis shows that these divergences correspond to distinct rhetorical phenomena: some probes capture discourse-level rhetorical stance embedded in extended argumentation, while others emphasize localized, syntax-driven interrogative acts. Together, these findings suggest that rhetorical questions in LLM representations are encoded by multiple linear directions emphasizing different cues, rather than a single shared direction.
\end{abstract}

\section{Introduction}

Rhetorical language is a common and important part of everyday communication. One of its most typical forms is the rhetorical question, which speakers use not to seek information, but to persuade, challenge, or signal stance. For example, the question “Do you really believe that?” is often used to express doubt rather than to elicit an answer, and “Who would ever agree with this?” functions as a critique rather than a genuine inquiry. In contrast to domains involving formal problem-solving or factual inquiry, which rely on literal interpretation, rhetorical questions convey non-literal meaning shaped by context, speaker intent, and discourse structure. As a result, understanding rhetorical questions provides an important perspective on how large language models interpret and generate language in real-world communicative settings.

While rhetorical and informational questions have been studied computationally for many years \citep{bhattasali2015automatic, oraby2017are, zhuang2020exploring, kikteva2024question}, how large language models internally represent rhetorical questions has received far less attention. Existing work on rhetorical question understanding has largely framed the problem as a question-answering or classification task with explicit labels \citep{ikumariegbe2025studying}.
While convenient, this formulation focuses on prediction accuracy, whereas our work takes a step forward in understanding the representational basis of rhetorical behavior within the model.

To achieve this, we adopt a perspective that examines how these models encode rhetorical question within their internal \emph{representational space}. Using multiple linear probes \citep{park2023linear}, we investigate where rhetorical signals emerge, whether probes learned within the same dataset capture similar structure, and how probes learned from different data sources compare and transfer across datasets. We find that rhetorical content is linearly separable from information-seeking questions within a given context and remains detectable under cross-dataset transfer. However, probing directions differ substantially within and across datasets. These differences reflect distinct rhetorical properties and lead to different rankings on the same corpus. \emph{This shows that strong probing accuracy or transferability does not imply that a property is captured by a single shared representational direction.} Instead, rhetorical questions are not organized along a single linear axis, but reflect multiple linguistic features that are emphasized differently depending on context and data. Our code is publicly available.\footnote{\url{https://github.com/ruyi101/rq-representation-probing}} We summarize our contributions as follows:
\begin{enumerate}
\item We present a systematic, data-driven analysis of rhetorical question in LLM representations across real-world contexts, showing that rhetorical signals emerge early and are most stably captured by last-token representations in decoder-only models.
\vspace{-2mm}
\item We reveal a divergence between discriminative performance and representational alignment: although rhetorical questions are consistently linearly separable and transferable, probes learned from different data distributions induce substantially different rankings with little overlap in the upper and lower ends of the ranking. 

\vspace{-2mm}
\item Through qualitative analysis, we demonstrate that rhetorical meaning is inherently heterogeneous, spanning discourse-level rhetorical stance and localized, syntax-driven interrogative acts rather than a single, unified representational dimension.
\end{enumerate}
\vspace{-2mm}


\section{Related Work}

\textbf{Rhetorical Questions.} Rhetorical questions have long been studied in linguistics and NLP, with early work focusing on their pragmatic and discourse functions \citep{jurafsky1997automatic, frank1990you, roberts1994people, han2002interpreting, vspago2016rhetorical}. More recent computational studies frame rhetorical questions as a classification or detection task using linguistic features and contextual cues \citep{bhattasali2015automatic, oraby2017are, zhuang2020exploring, kikteva2024question}. Recently, \citet{ikumariegbe2025studying} examined rhetorical questions in the context of large language models using QA-style judgments of whether a question in context is rhetorical or informational.  

Beyond rhetorical questions specifically, recent work has examined rhetorical behavior in LLMs at broader stylistic or strategic levels. \citet{qiu2025counterfactual} introduce a counterfactual framework for measuring rhetorical style independently of substantive content, and \citet{ji2025generalizable} develop LLM-based models for annotating rhetorical strategies across domains. In a different vein, \citet{reinhart2025llms} analyze grammatical and rhetorical variation in LLM-generated text at the level of model outputs. By contrast, our work focuses on rhetorical questions at the level of internal representations, examining how rhetorical intent is encoded, organized, and transferred across layers and datasets.  

Related work on sarcasm and irony \citep{zhang2025sarcasmbench, lee2025pragmatic} likewise addresses non-literal language, but centers on distinct pragmatic phenomena and does not directly examine the representation of rhetorical question intent in LLMs.

\noindent
\textbf{Interpretability of Representations.}
Linear probing, which evaluates whether a target property is linearly decodable from a model’s internal representations, has long been used to analyze neural networks \citep{alain2017understanding}. A broad line of work studies neural language models through interpretability and probing methods, using linear probes as diagnostic tools to assess whether linguistic or semantic properties are accessible from model representations. Recent work emphasizes population-level and training-free approaches, such as diffMean directions~\citep{marks2024the, vennemeyer2025sycophancy}, as well as sparse autoencoders~\citep{cunningham2024sparse, gao2025scaling, farnik2025jacobian, heap2025sparse} that linearly decompose activations into interpretable feature directions. Complementary geometric and information-theoretic analyses have also been applied to study representation spaces in LLMs \citep{hosseini2023large, skean2025layer}. Related work has also increasingly explored causal intervention-based methods \citep{meng2022locating, ghandeharioun2024patchscopes}, such as patching and model editing, to test how specific internal states contribute to downstream behavior. Our work builds on this representation-centric literature by applying linear probing and geometric analysis to rhetorical question intent across contexts.


\section{Methods}
In this section, we describe the datasets, representation choices, and linear probing framework used in our analysis.

\subsection{Datasets}

We conduct our analysis on two real-world rhetorical question datasets drawn from social media, which differ in domain, annotation protocol, and availability of contextual information.

\noindent\textbf{RQ.}
The RQ dataset introduced by \citet{zhuang2020exploring} consists of Twitter questions annotated as rhetorical or informational. The dataset contains 4,997 instances, with 2,332 labeled as rhetorical, and is split into 3,200 training, 797 validation, and 1,000 test samples. Each instance includes a target question and a prior tweet providing conversational context. In the main analysis, we focus on the \emph{question-with-context} formulation, which concatenates the prior tweet and the target question.
Under this formulation, instances average 38.9 tokens.\footnote{All token counts for both RQ and SRAQ use the Llama-3.3-70B-Instruct tokenizer.}

\noindent\textbf{SRAQ.}
The SRAQ dataset proposed by \citet{ikumariegbe2025studying} draws from Reddit conversation threads, where each example is built around a target question found within a single user's comment. Because a comment can span multiple paragraphs, the dataset provides context at different levels of granularity. We consider two: the \textit{Paragraph} formulation, which retains only the paragraph containing the target question, and the \textit{Full\_turn} formulation, which contains the entire comment. All instances are annotated as
rhetorical or informational.
The dataset includes 971 samples, split into 384 training, 103 validation, and 484 test instances, with 609 questions labeled as rhetorical. In the main paper, we use the \emph{Paragraph} formulation, which averages 58.5 tokens per instance. Results using the \emph{Full\_turn} formulation show similar trends and are reported in the Appendix~\ref{app:additional_results}.

Across both datasets, we use the original train-validation-test splits provided by the respective authors and do not modify the annotation labels.

\subsection{Representations}

Given an input sequence of tokens $\{x_1, \dots, x_T\}$, we extract hidden representations from a pretrained language model. Unless otherwise specified, we use the \emph{last-token representation} $h_T \in \mathbb{R}^d$, which is commonly used as a sequence-level summary in decoder-style models. 

To examine the effect of token aggregation, we also consider mean-pooled representations
\begin{equation}
    \bar{h} = \frac{1}{T} \sum_{t=1}^{T} h_t,
\end{equation}
where $h_t$ denotes the hidden state at token position $t$.

For fair comparison across probes, we primarily project representations into a PCA space with $k=64$ dimensions, defined separately for each dataset and model, and for each input formulation (with and without additional context). This choice is motivated by the widely adopted view that high-dimensional language model representations concentrate near a low-dimensional manifold, such that most task-relevant variation is captured by a relatively small number of principal components \citep{skean2025layer}. Projecting into a shared low-dimensional subspace reduces noise and stabilizes comparisons across probes while preserving the dominant structure of the representations.

Concretely, for a fixed dataset–model–input setting (e.g., paragraph vs.\ full\_turn), a single PCA transformation is applied to all examples, and all probes within that setting operate in the same projected space. Different datasets, models, or input formulations use different PCA projections. In addition to this shared PCA space, we also report selected results computed directly in the original embedding space, and verify that key diffMean trends remain similar (Appendix~\ref{app:pca}).

\subsection{Linear Probes}

We analyze rhetorical separability using three linear probes applied to fixed representations: a training-free population-level diffMean probe \citep{marks2024the}, and two trained discriminative probes based on logistic regression and linear support vector machine.

\noindent\textbf{DiffMean.}
The diffMean probe estimates a population-level direction by subtracting class-conditional means. Let $\mathcal{D}_+$ and $\mathcal{D}_-$ denote rhetorical and informational examples, respectively. The direction is
\begin{equation}
    w_{\text{DM}} = \mu_+ - \mu_-,
\end{equation}
where $\mu_\pm = \mathbb{E}_{x \in \mathcal{D}_\pm}[h(x)]$. 
Each example is scored by the inner product $w_{\text{DM}}^\top h(x)$. Larger values indicate stronger alignment of the representation $h(x)$ with the diffMean direction.

\noindent\textbf{Discriminative Linear Probes.}
We additionally train linear classifiers using logistic regression and hinge loss. Logistic regression learns a weight vector by minimizing the cross-entropy loss, which encourages probabilistic separation between rhetorical and informational examples. The hinge-loss probe, on the other hand, optimizes a margin-based objective corresponding to a linear support vector machine. In both cases, the learned weight vector $w$ defines a linear scoring function $w^\top h(x)$. Larger scores indicate stronger alignment with the learned separator, analogous to diffMean. The two probes differ only in their optimization objective while sharing the same linear hypothesis.

In our experiments, the diffMean direction is computed using the training split only, and the discriminative probes are trained on the training split and selected using validation performance. Unless otherwise noted, all results below are reported on the test split.

\subsection{Evaluation Metrics}

We use multiple evaluation metrics to characterize rhetorical separability and to compare the behavior of different linear probes. AUROC is used to evaluate each probe direction individually, measuring how well its scores separate rhetorical from informational questions on held-out data. To compare probes across datasets or objectives, we analyze both the similarity between probe directions and the agreement between their induced orderings.

\noindent\textbf{Rank Agreement.}
Let $s^{(a)}_i$ and $s^{(b)}_i$ denote the scores assigned to example $i$ by two probes (or by the same probe trained on two datasets). Ranking agreement is measured using the \emph{Spearman's rank correlation}:
\begin{equation}
\rho_s = \mathrm{corr}\!\left(\mathrm{rank}(s^{(a)}),\, \mathrm{rank}(s^{(b)})\right),
\end{equation}
where $\mathrm{rank}(\cdot)$ maps scores to ranks and $\mathrm{corr}(\cdot,\cdot)$ is the Pearson correlation applied to the rank vectors.

\noindent\textbf{Overlap at the tails.}
To assess agreement in the upper and lower ends of the ranking, for a fraction $p \in (0,1)$ we form the top-$p$ (or bottom-$p$) sets $A_p$ and $B_p$ under each scoring function, and compute the \emph{Jaccard index}:
\begin{equation}
J(A_p,B_p) = \frac{|A_p \cap B_p|}{|A_p \cup B_p|}.
\end{equation}
These metrics quantify whether different probes induce similar orderings globally ($\rho_s$) and whether they retrieve similar top- and bottom-ranked examples ($J$).


\section{Representation Choices for Rhetorical Probing}
\label{sec:representation-choices}

\begin{figure*}[ht!]
    \centering
    \includegraphics[width=\textwidth]{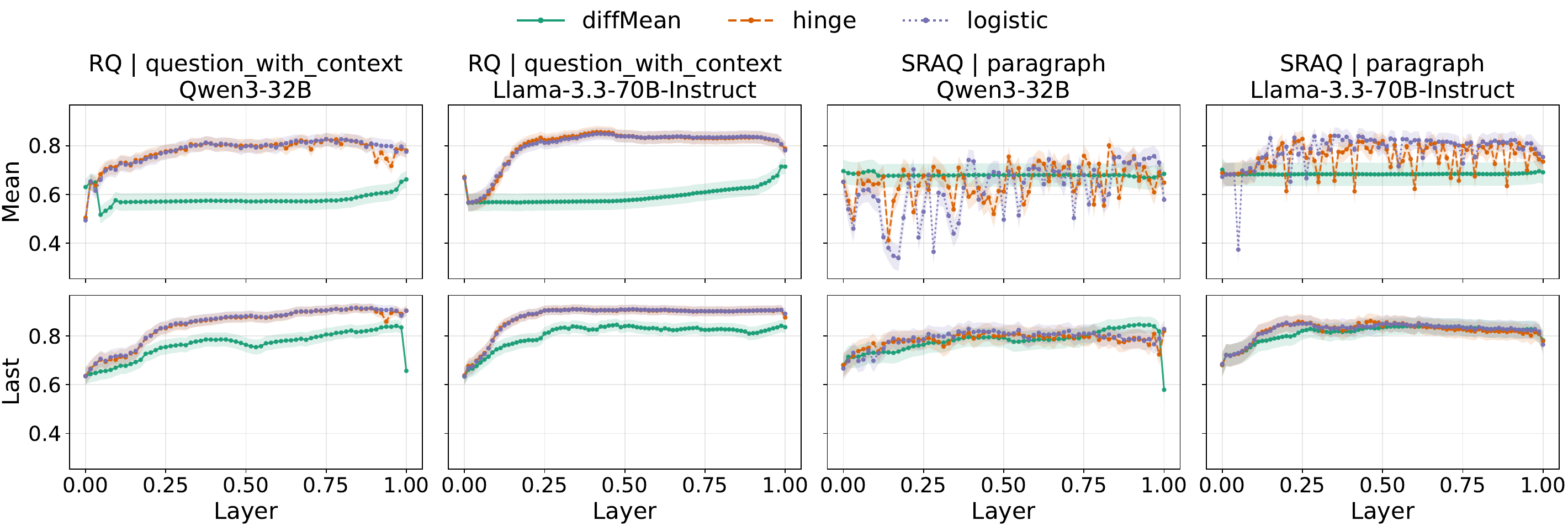}
    \caption{AUROC across layers and representations.
    Test AUROC across normalized layers using PCA-reduced representations. Rows compare mean-pooled and last-token embeddings; columns vary by dataset (RQ, SRAQ) and model (Qwen3-32B, Llama-3.3-70B).
    }
    \label{fig:layerwise-auroc}
\end{figure*}

We first examine how representation choice affects rhetorical probing in decoder-only models.\footnote{Our main analysis focuses on decoder-only models because they are the dominant setting for interactive language use. For completeness, we include a brief comparison with an encoder-based model in Appendix~\ref{app:bert}.} In the main text, we focus on two sequence-level representations: the final-token representation and mean pooling over all tokens. These choices reflect two contrasting assumptions about where rhetorical intent is encoded, namely whether it is concentrated near the end of the sequence or distributed across contextual cues. We therefore compare last-token and mean-pooled representations as our primary sequence-level views. For completeness, we also explore more fine-grained pooling variants, including pooling over the last few tokens and over the question span alone, and report these supplementary results in Appendix~\ref{app:different poolings}.

Figure~\ref{fig:layerwise-auroc} reports layer-wise test-set AUROC for rhetorical probing using PCA-reduced embeddings with 64 components, across two datasets (RQ and SRAQ) and two decoder-only models: Qwen3-32B~\citep{yang2025qwen3technicalreport} and Llama-3.3-70B~\citep{grattafiori2024llama3herdmodels}. Results are shown for mean-pooled and last-token representations. For each representation, we evaluate three linear probing directions: the diffMean vector, a hinge-loss classifier, and a logistic classifier.

\noindent
\textbf{DiffMean. }
For the diffMean direction, mean-pooled representations achieve AUROC comparable to last-token representations at early layers ($\approx0.60 - 0.65$). One plausible explanation is that mean pooling aggregates information from multiple tokens, making contextual signals available earlier, while last-token representations require deeper layers to accumulate similar context. At deeper layers, last-token representations generally achieve higher AUROC ($\approx 0.8$), suggesting that mean pooling is less effective because it aggregates token representations that are less informative for the rhetorical distinction.

\noindent
\textbf{Discriminative Probes. }
Hinge and logistic classifiers typically outperform diffMean across layers, with AUROC values around $0.85$--$0.9$ on RQ and $0.8$--$0.85$ on SRAQ for last-token representations, suggesting that learned linear decision boundaries better leverage the available representations. However, for both classifiers, mean-pooled representations do not provide an advantage over last-token representations. In particular, hinge and logistic probes applied to mean-pooled embeddings fail to surpass their counterparts operating on last-token embeddings, suggesting that, at the linear level with 64 PCA components, aggregating information across all tokens offers no additional benefit beyond the final token representation.

Dataset-dependent effects remain evident when using mean-pooled representations. While results on RQ remain relatively smooth across layers, performance on SRAQ is markedly more unstable across layers and across probing methods. This pattern suggests that, for more context-rich inputs, averaging across tokens can dilute rhetorical signals throughout the model.

Overall, mean pooling appears to retain useful lexical information at early layers, but its probing performance becomes noticeably less stable, particularly on SRAQ. Because mean-pooled representations neither outperform last-token representations under linear probes with 64 PCA components nor offer comparable stability, we focus the remainder of our analysis on last-token representations.

\section{Rhetorical Separability Across Linear Probes}
\label{sec:linear-separability}

Having established last-token representations as a stable choice, we next examine whether rhetorical questions are linearly separable and how different linear probes capture this structure.

\noindent
\textbf{AUROC. }
Figure~\ref{fig:layerwise-auroc} (second row) shows the AUROC achieved by different linear probes across layers and datasets. On RQ, hinge and logistic classifiers consistently achieve higher AUROC than the diffMean direction across a broad range of layers, with peak values approaching 0.9 at intermediate depths. This behavior is not surprising, as hinge and logistic probes are trained to optimize class separation, whereas diffMean is a training-free baseline. Across layers, hinge and logistic classifiers behave similarly, with no systematic difference in their peak performance.

In contrast, SRAQ shows a different pattern. All probes achieve AUROC well above chance. However, hinge and logistic classifiers provide only small improvements over the diffMean direction. Their performance closely tracks the training-free baseline across layers. Compared to RQ, the gap between trained probes and diffMean is much smaller.

Overall, AUROC results indicate that rhetorical questions are linearly separable from informational questions across models and datasets, though separability is not perfect. Appendix~\ref{app:steering} provides a steering analysis showing that perturbations along learned probing directions induce coherent changes in rhetorical behavior, consistent with these directions capturing a rhetorical signal.

\begin{figure*}[ht!]
    \centering
    \includegraphics[width=\textwidth]{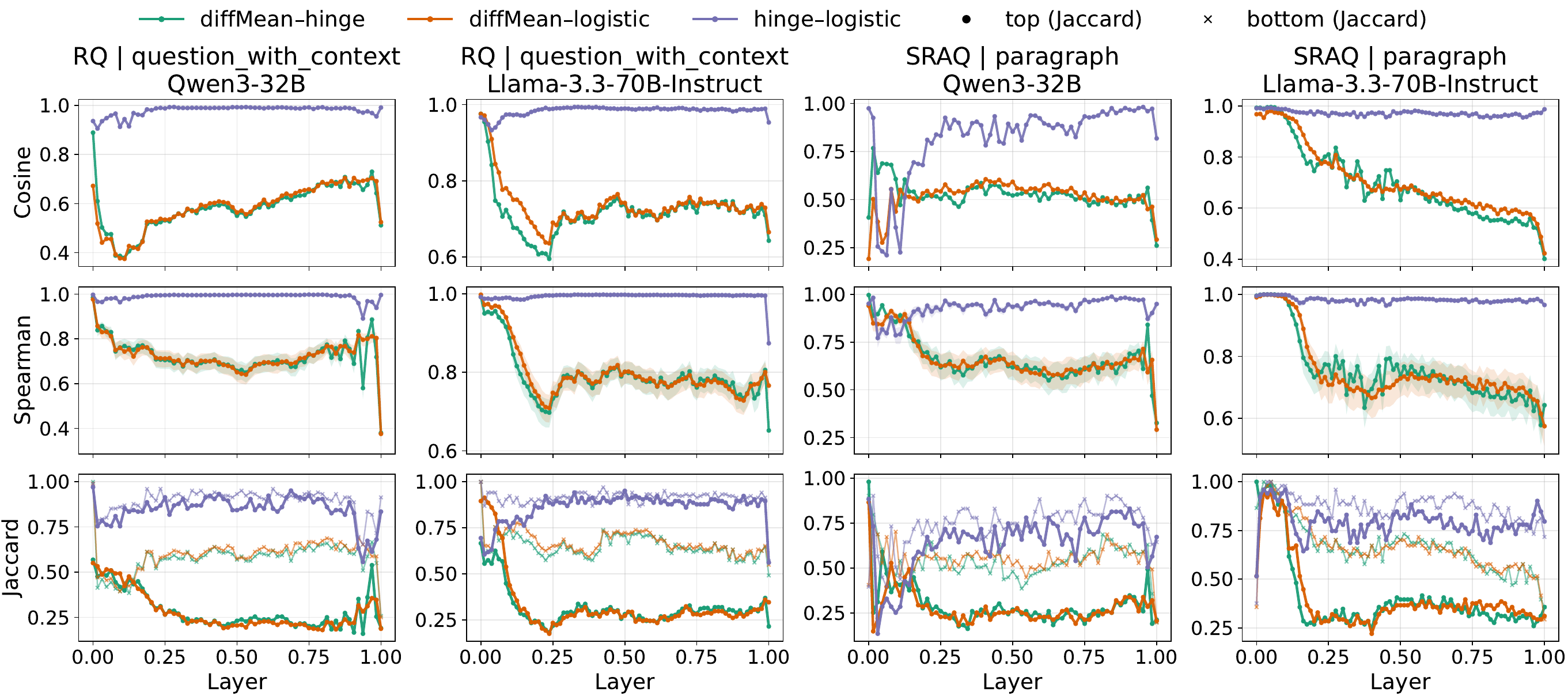}
    \caption{Alignment and ordering agreement between linear probes.
    Top: Cosine similarity between probing directions across layers.
    Middle: Spearman rank correlation between probe-induced scores, with shaded regions indicating confidence intervals.
    Bottom: Jaccard overlap between the top and bottom 20\% of examples ranked by each probe.
    Results are shown across datasets (RQ, SRAQ), models (Qwen3-32B, Llama-3.3-70B-Instruct), and layers.}

    \label{fig:probe-alignment}
\end{figure*}

\noindent
\textbf{Probing Alignment.} Now we examine the alignment between probing directions and the rankings they induce across layers to better understand the remaining gaps. 
Figure~\ref{fig:probe-alignment} shows the relationships among linear probing directions across layers, models, and datasets. Across all settings, hinge and logistic probes are nearly identical. Their cosine similarity remains close to $1$ across layers, with near-perfect Spearman correlation and high Jaccard overlap of top- and bottom-ranked examples. Despite different loss functions, both probes recover essentially the same linear direction and induce highly similar rankings.

In contrast, the relationship between diffMean and the trained probes is weaker and varies across datasets and layers. On RQ, diffMean shows moderate cosine similarity with hinge and logistic probes, mostly below $0.7$, especially at intermediate and later layers. Spearman correlations range between $0.6$ and $0.8$, indicating only partial agreement in the induced rankings. This alignment is consistently lower and more layer-dependent than that observed between hinge and logistic probes.

On SRAQ, divergence is stronger: cosine similarity between diffMean and trained probes is lower and less stable ($\approx 0.5$), with reduced rank agreement (Spearman $\approx 0.6$ at middle to late layers).

This pattern is also reflected in the Jaccard overlap of highest- and lowest-ranked examples, shown in the third row of Figure~\ref{fig:probe-alignment}. Across models and datasets, overlap among top-ranked examples is low, around $0.25$, while overlap among bottom-ranked examples is higher, typically around $0.5$. This asymmetry suggests that informational instances are ranked more consistently across probes, whereas highly rhetorical instances show greater variability. In this sense, rhetorical status appears more heterogeneous than informational status.

This asymmetry helps explain the AUROC results on SRAQ. Probes with similar AUROC can induce divergent rankings and little overlap in their top- and bottom-ranked examples, hinting that comparable discriminative performance does not necessarily imply shared representational properties. Instead, probes with similar AUROC may capture different aspects of rhetorical signal.


\section{Transferability of Rhetorical Separators}

So far, our analysis has focused on within-dataset settings, where probe directions are learned and evaluated on the same data distribution. If rhetorical intent corresponded to a robust linear structure in representation space, such structure should generalize under distributional shift. We therefore examine cross-dataset transfer by learning probe directions on one dataset and applying them to representations from the other. We evaluate both separability and agreement relative to probes learned directly on the target dataset. Because probes are learned in dataset-specific PCA subspaces, we map directions back to the full embedding space prior to comparison (see Appendix~\ref{app:pca-mapback}).

\begin{figure*}[ht]
    \centering
    \includegraphics[width=0.96\textwidth]{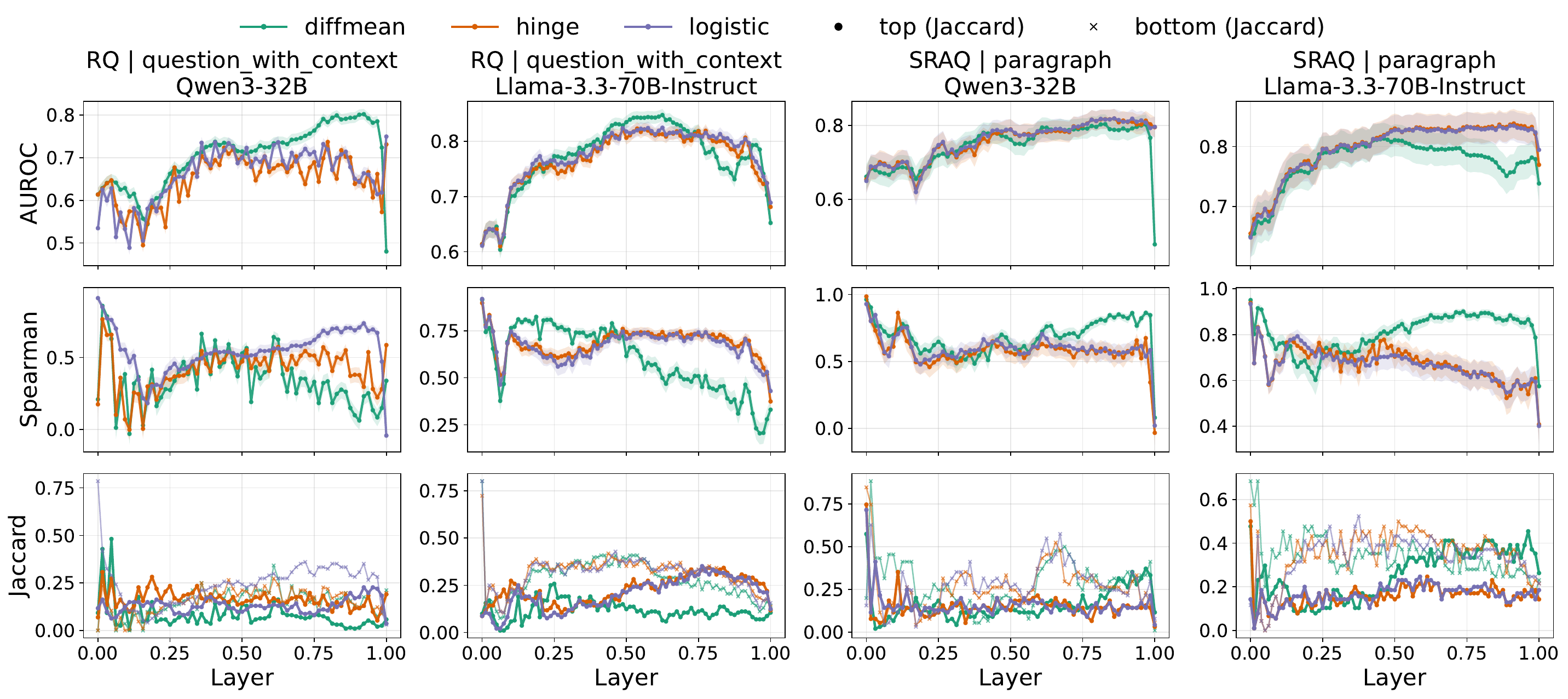}
    \caption{
    Transferability of rhetorical probing directions across datasets.
    For RQ panels (left), directions are learned on SRAQ and applied to RQ; for SRAQ panels (right), directions are learned on RQ and applied to SRAQ.
    Rows report test AUROC of transferred directions (top), Spearman rank correlation between rankings induced by transferred directions and rankings induced by directions learned on the target dataset (middle), and Jaccard overlap between the top and bottom 20\% of examples under these two rankings (bottom).
    }
    \label{fig:probe-transferability}
\end{figure*}

\noindent
\textbf{AUROC under transfer.} 
We first examine cross-dataset transfer in terms of discriminative performance. As shown in the first row of Figure~\ref{fig:probe-transferability}, probe directions learned on one dataset exhibit a modest drop in AUROC when applied to the other, but still achieve values around $0.7$–$0.8$ across models and layers. This indicates that rhetorical intent contains a partially shared linear component across datasets.

\noindent
\textbf{Ranking agreement. }
In contrast to AUROC, ranking agreement under transfer is much weaker. For each target dataset, we score the same examples using both the in-domain direction and the transferred direction, rank the examples by their projection scores, and compute Spearman correlation between the two resulting rankings. As shown in the second row of Figure~\ref{fig:probe-transferability}, these correlations are only moderate overall. For trained probes, they often fall toward $0.5$ or lower at deeper layers. The diffMean directions follow a similar trend, with some deviations across datasets and layers, but they likewise do not show strong agreement. This shows that transferability in AUROC does not imply close alignment in the induced rankings.

Agreement among top- and bottom-ranked examples deteriorates further under transfer. As shown in the third row of Figure~\ref{fig:probe-transferability}, Jaccard overlap between the top and bottom $20\%$ of examples is consistently low, particularly for top-ranked instances, where it often falls below $0.2$. Overlap among bottom-ranked examples is higher but remains limited, around $0.3$–$0.4$. As in the within-dataset setting, this asymmetry suggests that informational instances are identified more consistently across datasets, whereas highly rhetorical instances are more heterogeneous and dataset-dependent.

\begin{figure}[t]
    \centering
    \includegraphics[width=0.96\linewidth]{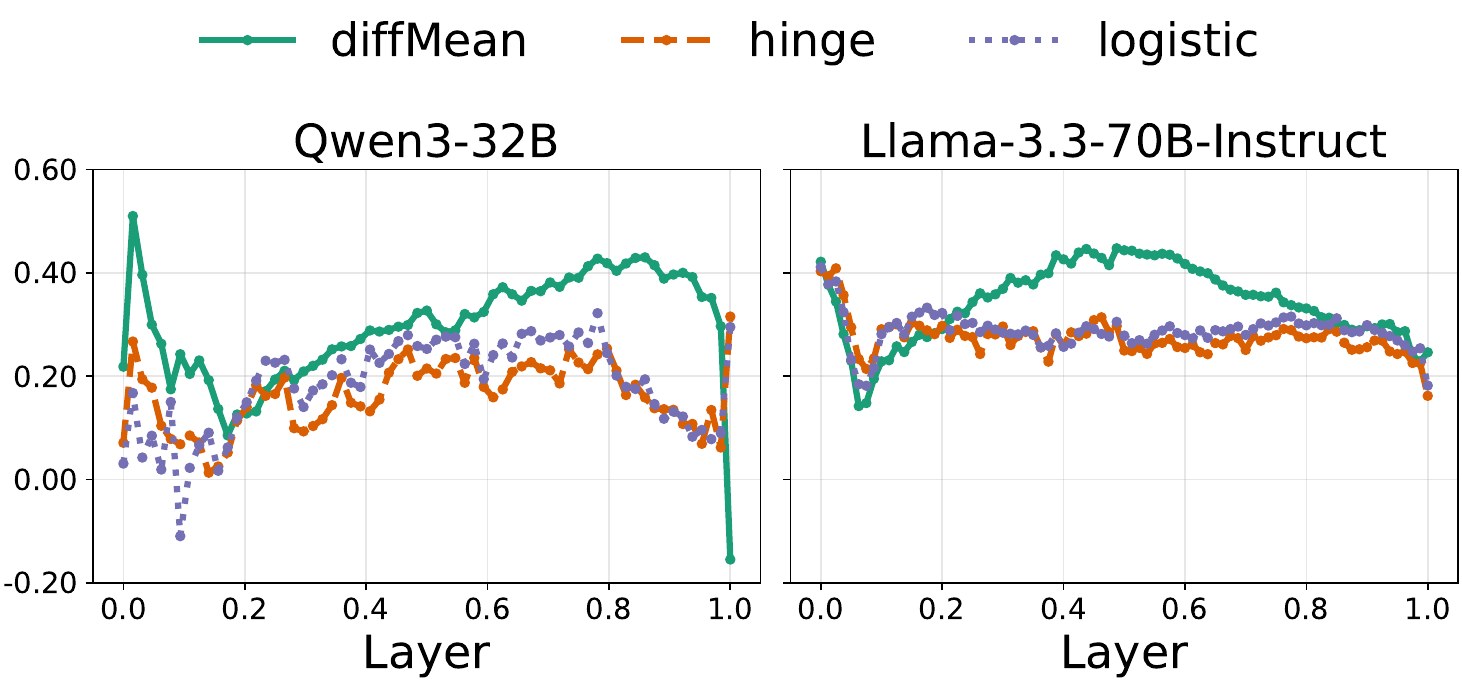}
    \caption{Cross-dataset alignment of rhetorical probing directions.
    Cosine similarity between probing directions learned on RQ and SRAQ, measured layer-wise for two models and three different probings.}
    \label{fig:cross-dataset-cosine}
\end{figure}

\noindent
\textbf{Directional alignment. }
To better understand these transfer behaviors, we analyze \emph{directional alignment under transfer}. Figure~\ref{fig:cross-dataset-cosine} reports cosine similarity between probe directions learned on RQ and SRAQ across layers. Across models, all probes show low similarity, typically around $0.2$--$0.4$, indicating that directions learned on different datasets are not collinear. Hinge and logistic probes exhibit relatively stable alignment across layers. In contrast, the diffMean direction is more variable, reaching higher similarity at intermediate layers but remaining limited to around $0.4$ for Llama-3.3-70B-Instruct and at later layers for Qwen3-32B, with lower alignment elsewhere.

The transfer results provide clear evidence about the structure of rhetorical questions in representation space. Under distributional shift, probe directions retain meaningful separability while showing limited alignment, moderate rank agreement, and low overlap at the rank ends. Together, these observations suggest that, although rhetorical signal is present, it is not organized along a single linear direction. Instead, rhetorical questions appear to be expressed heterogeneously, with different non-collinear directions capturing distinct aspects of rhetorical usage. Similar patterns in the within-dataset analyses support this interpretation.


\section{Qualitative Insights into Rhetorical Probing}
\label{sec:qualitative}

\begin{table*}[t]
\centering
\small
\setlength{\tabcolsep}{6pt}
\renewcommand{\arraystretch}{1.3}

\begin{tabularx}{0.98\textwidth}{@{} >{\centering\arraybackslash}p{0.9cm} >{\centering\arraybackslash}p{0.9cm} >{\RaggedRight\arraybackslash}X >{\centering\arraybackslash}p{1.8cm} @{}}
\toprule
\makecell[c]{\textbf{SRAQ}\\\textbf{rank}} &
\makecell[c]{\textbf{RQ}\\\textbf{rank}} &
\textbf{Paragraph} &
\textbf{Gold Label} \\
\midrule

1 & 15 &
\emph{I find our infatuation with ourselves to be a bit self serving.
\textbf{Who but us cares?}
It's in our own self interest to think we're great.
But \textbf{what does it accomplish?} Nothing.
\dots We think we're so great because we can manipulate things.
But more often than not this manipulation simply creates major
problems \dots
``We can think of magnificent things; just look at the incredible
architecture we've created.''
\textbf{So?} Spiders spin incredible webs. Birds weave beautiful nests.
Bugs create towers that they live in.
Complexity does not necessarily mean superiority.}
& rhetorical \\

\addlinespace[0.6em]

2 & 108 &
\emph{\dots What anti-copyright fanfic enthusiasts like yourself don't
realize is that you're just trying to have your cake and eat it too.
\dots You can do that already with public domain material or with
your own distilled characters and settings. But you don't want to.
\textbf{Why not?} Because you REALLY WANT the popularity of these
modern franchises to hold you up and make your story make sense.
You're not asking for the ideas, you already have them. You're
asking for the popularity and depth of concept; and nobody promised
you that.}
& rhetorical \\

\addlinespace[0.6em]

3 & 240 &
\emph{\dots If you scrutinise the majority of Reykjavik sized
``hellhole'' cities, you will find that they are often an extension
of a greater metropolitan area. For example, Watford is the same
size as Reykjavik and Exeter is the same size as Reykjavik.
Exeter has a lower crime rate than Watford ---
\textbf{why?} Because Watford is essentially a city within the
greater metropolitan area of London.}
& rhetorical \\

\midrule

230 & 1 &
\emph{My guess is you'll have to go to or phone their office to
apply \dots In most states the initial court appearance is considered
a ``critical stage of the process'' where you are entitled to
Counsel \dots If this is your first offense most places have a
pre-trial diversion program \dots Talk to your lawyer \dots
Whatever happens this will still likely be expensive.
A ``social thing.''
\textbf{What? you only drink to make other people more
interesting?} :)}
& rhetorical \\

\addlinespace[0.6em]

146 & 2 &
\emph{\dots so \textbf{why would they even try} and be granted legal
visitation when they know the issue will be looked at in court?
Are they trying to somehow go around this issue? And if so, how?
\dots They just closed the suit with his first wife with 3 kids
and that one lasted since 1994--2017 \dots
It's not for custody, they are asking for visitation rights,
which is ridiculous.}
& informational \\

\addlinespace[0.6em]

237 & 3 &
\emph{\textbf{Wait, what?} Not a single higher spot in Estonia,
but a few bumps in Latvia and Lithuania. Do your research, we
have huge ``mountains'' compared to them \textup{=)}}
& rhetorical \\

\bottomrule
\end{tabularx}

\caption{SRAQ instances ranked by inner product with
the SRAQ-derived diffMean direction (top) and RQ-derived
(bottom) at layer 48 of Qwen3-32B. Each row reports both
ranks. Bold text marks the target question; ellipses indicate
truncated text for readability.\protect\footnotemark}
\label{tab:qual-top3-refined}
\end{table*}

To move beyond aggregate performance metrics, we conduct a qualitative analysis of the rankings induced by diffMean directions at layer 48 of Qwen3-32B. We compute one diffMean direction from SRAQ and one from RQ, then score every SRAQ instance by the inner product of its representation with both directions respectively ($w_{\text{DM}}^\top h(x)$; see Section~3.3). For each direction, we rank all SRAQ instances by this score and retain the three highest-ranked examples. Table~\ref{tab:qual-top3-refined} shows the results: the top rows list the three examples ranked highest by the SRAQ-derived direction, and the bottom rows list those ranked highest by the RQ-derived direction. Each row also reports the same example's rank under the other direction.

\footnotetext{Some top-ranked instances are omitted here because they contain socially sensitive content; the complete rankings are reproducible from the released code.}

The top rows of Table~\ref{tab:qual-top3-refined} show that the SRAQ-derived direction prioritizes passages in which
rhetorical questions serve as structural scaffolding for
extended arguments. In the top-ranked example, three
successive questions (``Who but us cares?'', ``what does it
accomplish?'', ``So?'') each open a new stage of a
philosophical argument about human self-importance; in the
second, ``Why not?'' sets up a multi-sentence explanation
of why fanfiction writers seek borrowed popularity rather
than building their own. In each case, the rhetorical
question drives the discourse forward and organizes the surrounding argument.

In contrast, the bottom rows show that the RQ-derived
direction favors short, syntax-driven interrogative forms
whose rhetorical force is localized. The top-ranked
example is a paragraph of legal advice in which a
rhetorical question appears only as a throwaway joke in the
final sentence; the third-ranked instance is a two-sentence
expression of surprise. Most surprisingly, the second-ranked
example is labeled \emph{informational} in the gold
annotations: it contains genuine requests for clarification
(``why would they even try\dots?'', ``Are they trying to
somehow go around this issue?'') that the RQ-derived
direction ranks highly because of their surface
interrogative form, not because of any rhetorical intent.

To make this qualitative contrast more concrete, we consider input length as a simple quantitative proxy. While discourse-level properties such as stance-taking or counterargument structure are difficult to measure reliably without additional annotation, length provides a lightweight way to test whether the two directions emphasize different types of examples. As shown in Table~\ref{tab:length_top_ranked}, under the same ranking setup the top-ranked SRAQ examples selected by the in-domain SRAQ diffMean direction are substantially longer on average than those selected by the transferred RQ diffMean direction. This pattern is consistent with our interpretation that the SRAQ direction more often captures broader discourse context, whereas the RQ direction more often emphasizes more localized rhetorical cues.

\begin{table}[t]
\centering
\resizebox{0.9\linewidth}{!}{
\begin{tabular}{lcc}
\toprule
Subset & SRAQ direction & RQ direction \\
\midrule
Top 1\% & 188.8 & 126.4 \\
Top 3\% & 150.9 & 116.4 \\
\bottomrule
\end{tabular}
}
\caption{Mean token length of top-ranked SRAQ examples selected by each diffMean direction under the ranking setup used in Section~7. Token counts are computed using the Llama-3.3-70B-Instruct tokenizer.}
\label{tab:length_top_ranked}
\end{table}

Ultimately, this evidence suggests that rhetorical meaning does not manifest as a single dominant direction, but rather as a set of \emph{distinct, emergent rhetorical properties}. These properties span different rhetorical functions, ranging from localized discourse repair to global rhetorical stance, indicating that rhetoric in LLMs is inherently heterogeneous and context-sensitive.


\section{Conclusion and Future Work}
In this work, we study how rhetorical questions are encoded in large language models using linear probes across two social media datasets. We find that rhetorical content is reliably linearly separable using a single embedding for each input sequence, and that last-token representations provide more stable signals than mean pooling. Although probing directions transfer across datasets, discriminative performance and representational alignment do not always coincide. Probes with similar AUROC can induce substantially different rankings on the same data, with little overlap in the top- and bottom-ranked examples, indicating that similar accuracy can reflect different underlying representations. Qualitative analysis suggests that this divergence reflects the heterogeneous nature of rhetorical questions, which range from discourse-level stance-taking in extended arguments to localized, turn-level interrogative acts.

More broadly, these results caution against treating strong probing performance or successful cross-dataset transfer as evidence of a single shared representational dimension. Linear probes can instead recover different directions that are all effective for discrimination but are not aligned with one another. This suggests that rhetorical questions are encoded in a context-sensitive and heterogeneous manner, rather than along a single linear feature. Future work should clarify how representational features associated with individual directions can be defined and validated, and how such features can be distinguished from collections of directions that perform similarly but capture different aspects of the phenomenon.

Another important direction is to study whether the rhetorical signals identified here are not only encoded, but also controllable. Although we find that rhetorical intent is linearly separable in representation space and our preliminary steering experiments suggest that the identified directions can induce changes in rhetorical behavior, we emphasize that linear separability does not itself imply linear controllability. Extending this analysis with more systematic causal interventions is an important direction for future work.

\section*{Limitations}
Our empirical analysis is restricted to two datasets drawn from social media domains. While other datasets related to rhetorical or discourse phenomena are available, they generally lack labeling of comparable reliability, consistency, or granularity, which limits their suitability for the type of fine-grained probing analysis conducted in this work. Moreover, the datasets considered here are drawn from real-world data and therefore exhibit inherent noise, which may further obscure or attenuate underlying signals. For these reasons, we limit our evaluation to these datasets, and our findings should be interpreted within this scope.

In addition, although we analyze representation dynamics across layers, our methodology focuses exclusively on linear probing of representation spaces. This choice emphasizes interpretability and analytical clarity, but it necessarily excludes signals that may be encoded through nonlinear interactions, alternative activation pathways, or mechanisms that are not linearly separable in the representation space considered here.

\section*{Acknowledgements}
This work is supported in part by a URC Faculty Scholars Research Award from the Office of Research at the University of Cincinnati. We thank the CincyNLP group for helpful discussions, and the anonymous reviewers for their valuable feedback. LHY would like to thank Xintong Cui for pointing out a notation error in the appendix.

\nocite{panickssery2024steering,turner2024steering}
\bibliography{ref}

\appendix

\newpage
\section{Effect of PCA Truncation on Linear Probing}
\label{app:pca}
\begin{figure}[ht!]
    \centering
    \begin{subfigure}[t]{0.95\linewidth}
        \centering
        \includegraphics[width=\linewidth]{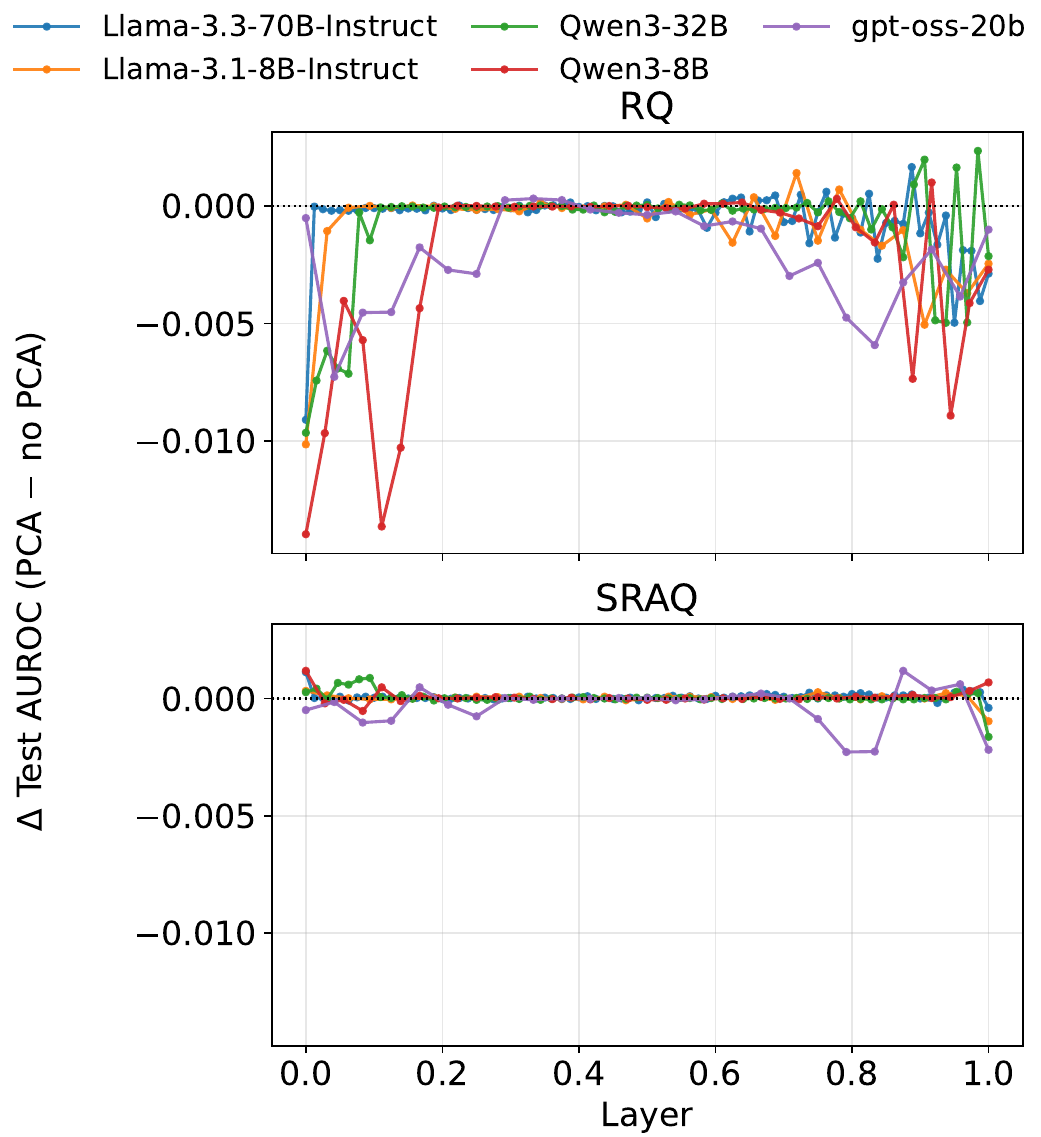}
        \caption{Mean token.}
        \label{fig:auroc_pca_minus_nopca_mean}
    \end{subfigure}

    \vspace{0.8em}

    \begin{subfigure}[t]{0.95\linewidth}
        \centering
        \includegraphics[width=\linewidth]{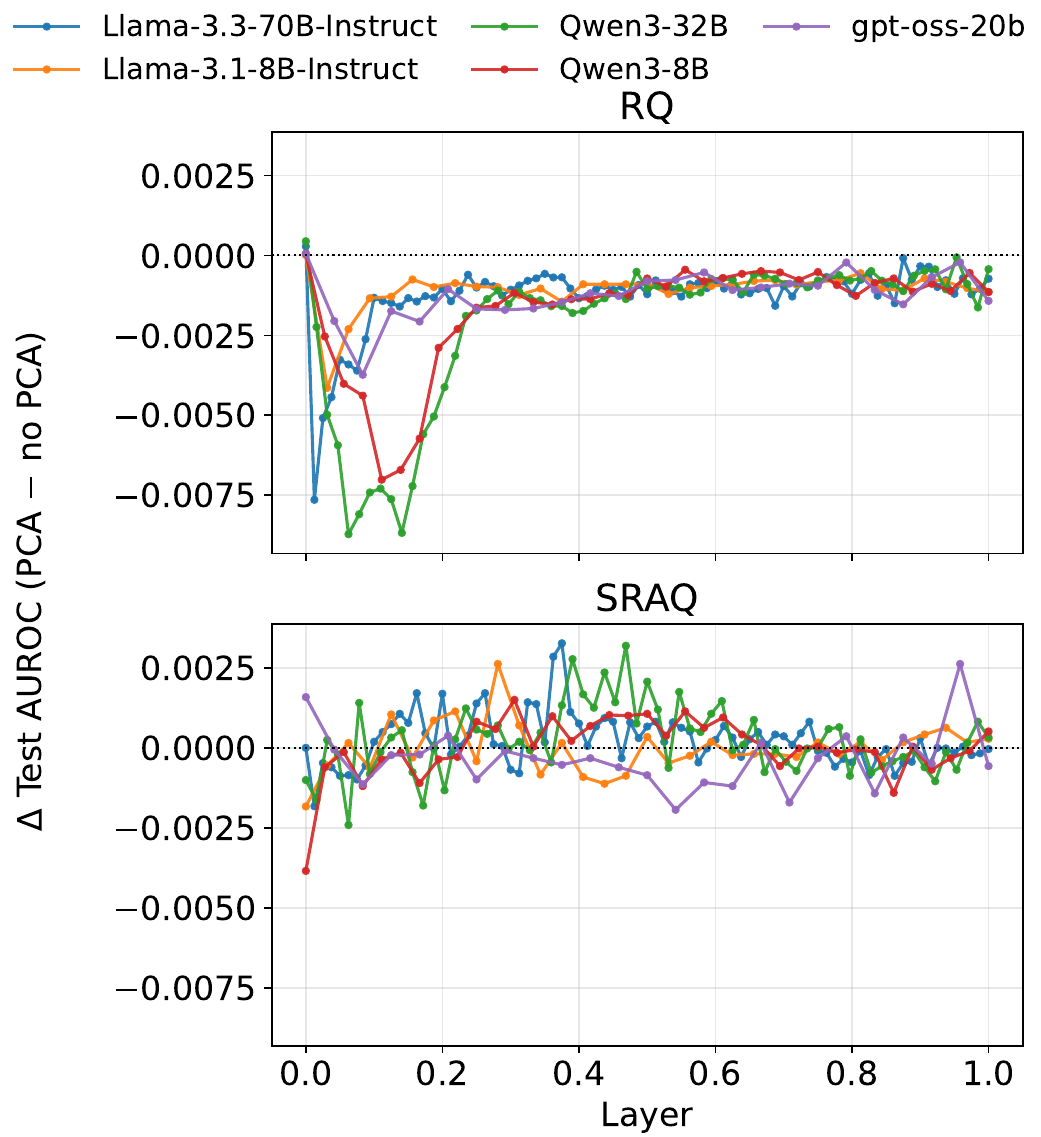}
        \caption{Last token.}
        \label{fig:auroc_pca_minus_nopca_last}
    \end{subfigure}

    \caption{AUROC differences (PCA minus no PCA) for diffMean linear probes under mean-token and last-token pooling.}
    \label{fig:auroc_pca_minus_nopca_tokens}
\end{figure}

In this section, we compare diffMean probe performance using the first 64 PCA components versus the full embedding spaces. As shown in Figure~\ref{fig:auroc_pca_minus_nopca_tokens}, the resulting AUROC differences are consistently small across models and layers, remaining below one percentage point in magnitude. This indicates that, for diffMean probes, projecting representations onto a low-dimensional PCA subspace does not meaningfully alter discriminative performance relative to using the full embedding space.

\begin{figure}[ht!]
    \centering
    \begin{subfigure}[t]{0.95\linewidth}
        \centering
        \includegraphics[width=\linewidth]{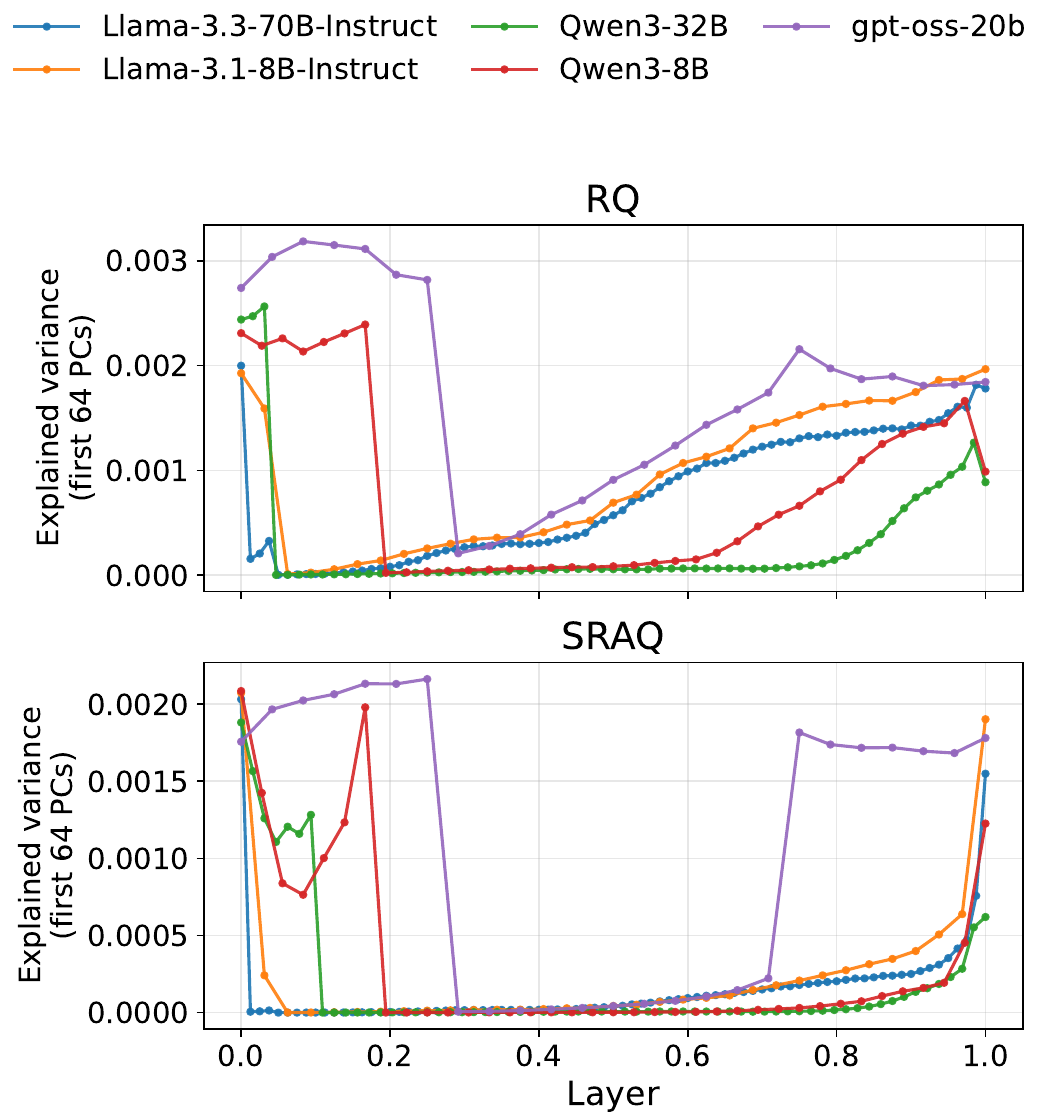}
        \caption{Mean-token representations.}
        \label{fig:pca64_explained_mean}
    \end{subfigure}

    \vspace{0.6em}

    \begin{subfigure}[t]{0.95\linewidth}
        \centering
        \includegraphics[width=\linewidth]{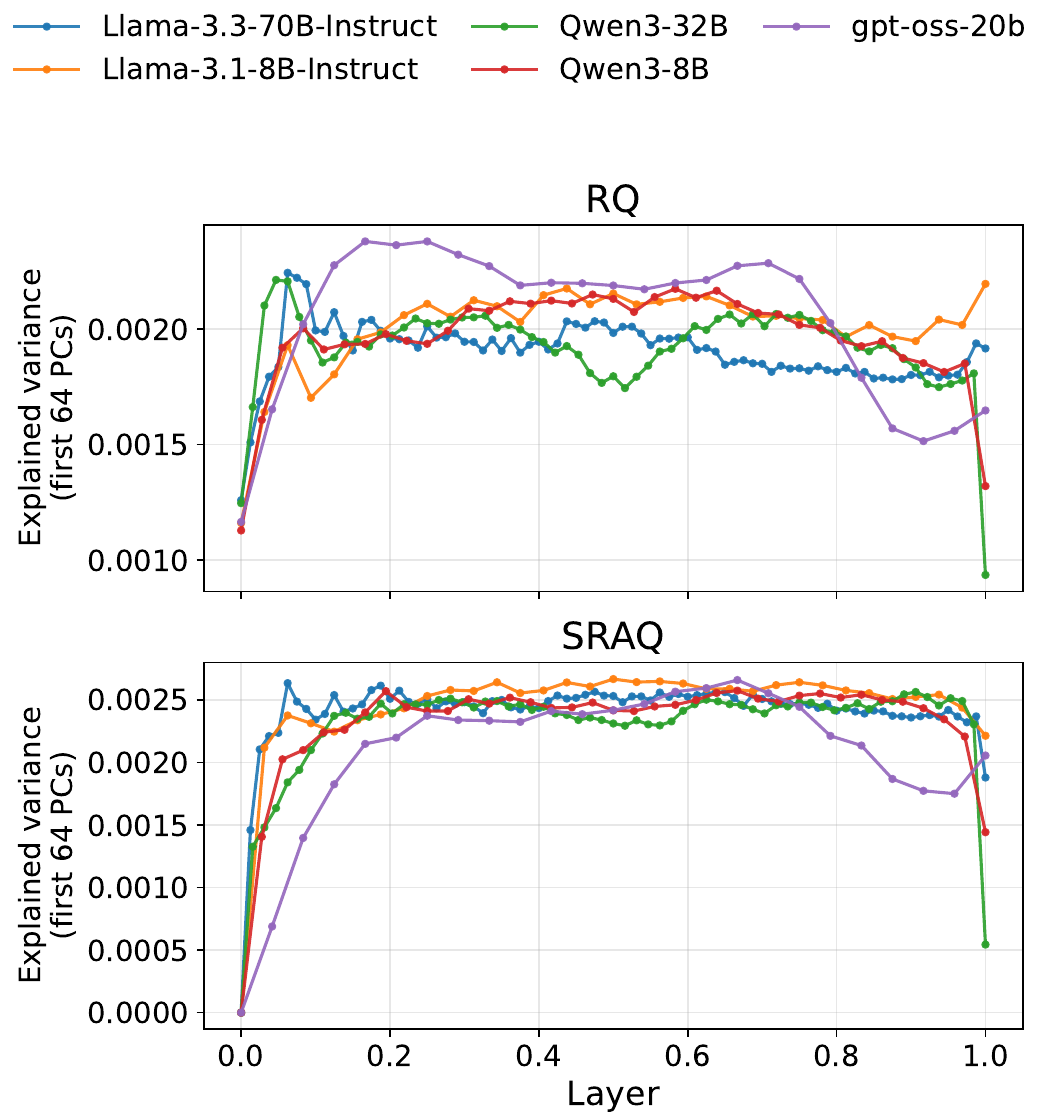}
        \caption{Last-token representations.}
        \label{fig:pca64_explained_last}
    \end{subfigure}

    \caption{Per-layer explained-variance ratio of the 64th principal component (i.e., the marginal variance explained by the 64th principal component) for mean-token vs.\ last-token representations.}
    \label{fig:pca64_explained_ratio}
\end{figure}

To further contextualize this result, Figure~\ref{fig:pca64_explained_ratio} reports the per-layer explained-variance ratio of the 64th principal component for both mean-token and last-token representations. Across models and layers, the variance explained by the 64th component remains well below 1\%, indicating that even the highest-index component retained in our PCA truncation captures only a very small fraction of the total variance. Consequently, all subsequent components beyond the 64th are expected to contribute even less. This provides a geometric explanation for the negligible AUROC differences observed above: restricting representations to the leading 64 principal components preserves nearly all variance relevant to diffMean linear probes, while discarding directions that collectively account for only a minor fraction of the embedding space.

This observation motivates our choice to restrict representations to the first 64 principal components in subsequent experiments involving other linear probes, including logistic and hinge classifiers. Since higher-index components explain only a negligible fraction of the total variance, retaining them is unlikely to add meaningful discriminative signal and may instead introduce noise or numerical instability. In practice, projecting onto the leading 64 components yields more stable training and evaluation behavior for these probes without materially affecting performance. We therefore focus on this dimensionality throughout the remainder of our linear probing analyses.

\section{Mapping Linear Probes from PCA to Embedding Space}
\label{app:pca-mapback}

In our main experiments, logistic-regression and hinge-loss probes are trained in a PCA-reduced space of dimension $k$ (here $k{=}64$) for numerical stability and consistent comparisons across probes. Because PCA projections differ across datasets, expressing a learned probe in the original embedding space of dimension~$d$ allows for consistent interpretation and comparison.

\noindent\textbf{PCA projection.}
Let $x \in \mathbb{R}^{d}$ denote an original (sequence-level) representation, and let $\mu \in \mathbb{R}^{d}$ be the PCA mean computed on the training split for a fixed dataset--model--input setting. Let $W \in \mathbb{R}^{k \times d}$ be the PCA loading matrix whose rows are the top-$k$ principal directions (orthonormal). The PCA coordinates are
\begin{equation}
z \;=\; W (x - \mu) \in \mathbb{R}^{k}. 
\end{equation}

\noindent\textbf{Linear probe in PCA space.}
Both logistic regression and hinge-loss (linear SVM) define an affine scoring function in PCA space,
\begin{equation}
s(z) \;=\; w_z^\top z + b,
\end{equation}
where $w_z \in \mathbb{R}^{k}$ is the learned weight vector and $b \in \mathbb{R}$ is the bias/intercept.

\noindent\textbf{Mapping weights back to the original space.}
Substituting $z = W(x-\mu)$ yields
\begin{align}
s(x)
&= w_z^\top W (x-\mu) + b \\
&= (W^\top w_z)^\top x \;+\; \bigl(b - (W^\top w_z)^\top \mu \bigr).
\end{align}
Therefore, the equivalent probe in the original embedding space has weight vector
\begin{equation}
w_x \;=\; W^\top w_z \in \mathbb{R}^{d}
\end{equation}
and bias
\begin{equation}
b_x \;=\; b - w_x^\top \mu.
\end{equation}
This mapping preserves scores exactly for any $x$ under the same PCA transform. Intuitively, the mapped-back classifier places all its mass in the PCA subspace; components orthogonal to $\mathrm{span}(W)$ have zero weight.

\section{Geometric Characterization of Dataset Differences}
\label{app:subspace-alignment}
Throughout the paper, we observe that RQ and SRAQ exhibit systematically different rhetorical behaviors, both in qualitative examples and in the linear directions identified by probing.
Motivated by these behavioral differences, we further examine whether they are reflected in the geometry of the representation spaces already analyzed in the main text.

Specifically, we reuse the same PCA-based subspace construction employed throughout the paper.
For each model and layer, we form a $64$-dimensional subspace using the top-$64$ principal components of the embedding distributions induced by each dataset.
This allows us to directly compare the geometric structure of the dominant variance directions underlying the observed rhetorical signals, without introducing a new representation.

We compare the resulting subspaces across datasets using two complementary alignment measures, which capture different geometric aspects.
The first is the geodesic distance between subspaces on the Grassmann manifold, computed from their principal angles.
This metric depends only on the subspaces themselves and is invariant to the ordering or orientation of individual basis vectors.
As a result, it provides a global measure of subspace alignment, quantifying how similarly the two datasets organize their dominant variance directions as a whole.

The second measure is the mean cosine similarity between corresponding principal components.
Unlike geodesic distance, this metric is sensitive to the alignment of individual PCA directions and implicitly depends on their ordering.
It therefore captures whether not only the subspaces overlap, but also whether their leading directions of variation are aligned in a consistent manner.
In practice, the mean cosine similarity remains close to zero across layers, indicating that even when subspaces are not maximally separated, their dominant directions are largely misaligned.

\begin{figure}
    \centering
    \includegraphics[width=0.45\textwidth]{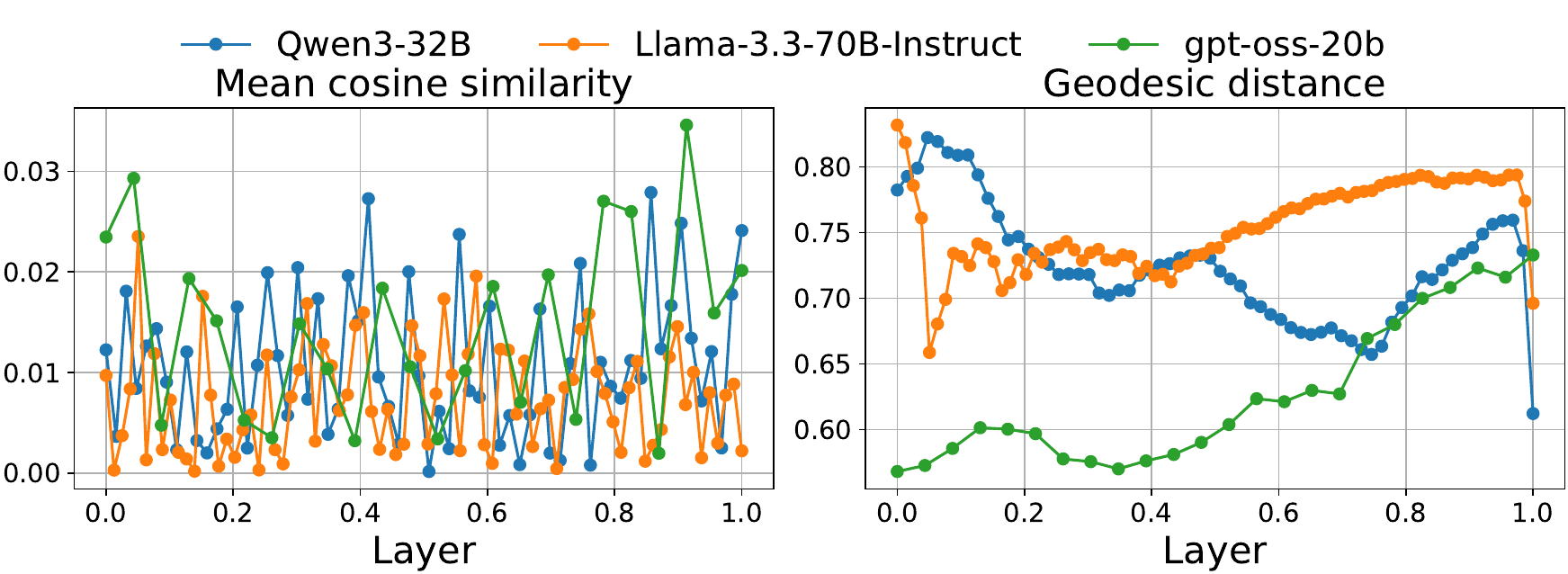}
    \caption{Subspace alignment between RQ and SRAQ across layers for multiple models. We report cosine similarity and geodesic distance between the corresponding layer-wise subspaces, using normalized layer index on the $x$-axis. Higher distances indicate weaker cross-dataset alignment.}
    \label{fig:subspace-alignment-data}
\end{figure}

Figure~\ref{fig:subspace-alignment-data} reports both metrics across normalized layer depth for multiple models.
Together, these measurements provide a geometric characterization of dataset differences from two perspectives: overall subspace alignment, and fine-grained alignment of leading variance directions, complementing the behavioral differences observed throughout the paper.

In addition, we observe that \texttt{GPT-OSS-20B}~\citep{openai2025gptoss120bgptoss20bmodel} exhibits qualitatively different behavior from the other models considered.
Across layers, this model shows distinct trends in geodesic distance, a pattern that is consistent with its behavior observed under multiple other analyses in the paper.
While we do not attribute this difference to a specific architectural or training factor, the consistency of this effect across independent metrics suggests that it reflects a systematic property of the model’s representations rather than measurement noise.

\section{DiffMean Probing with Alternative Pooling}
\label{app:different poolings}
\begin{figure}
    \centering
    \includegraphics[width=0.48\textwidth]{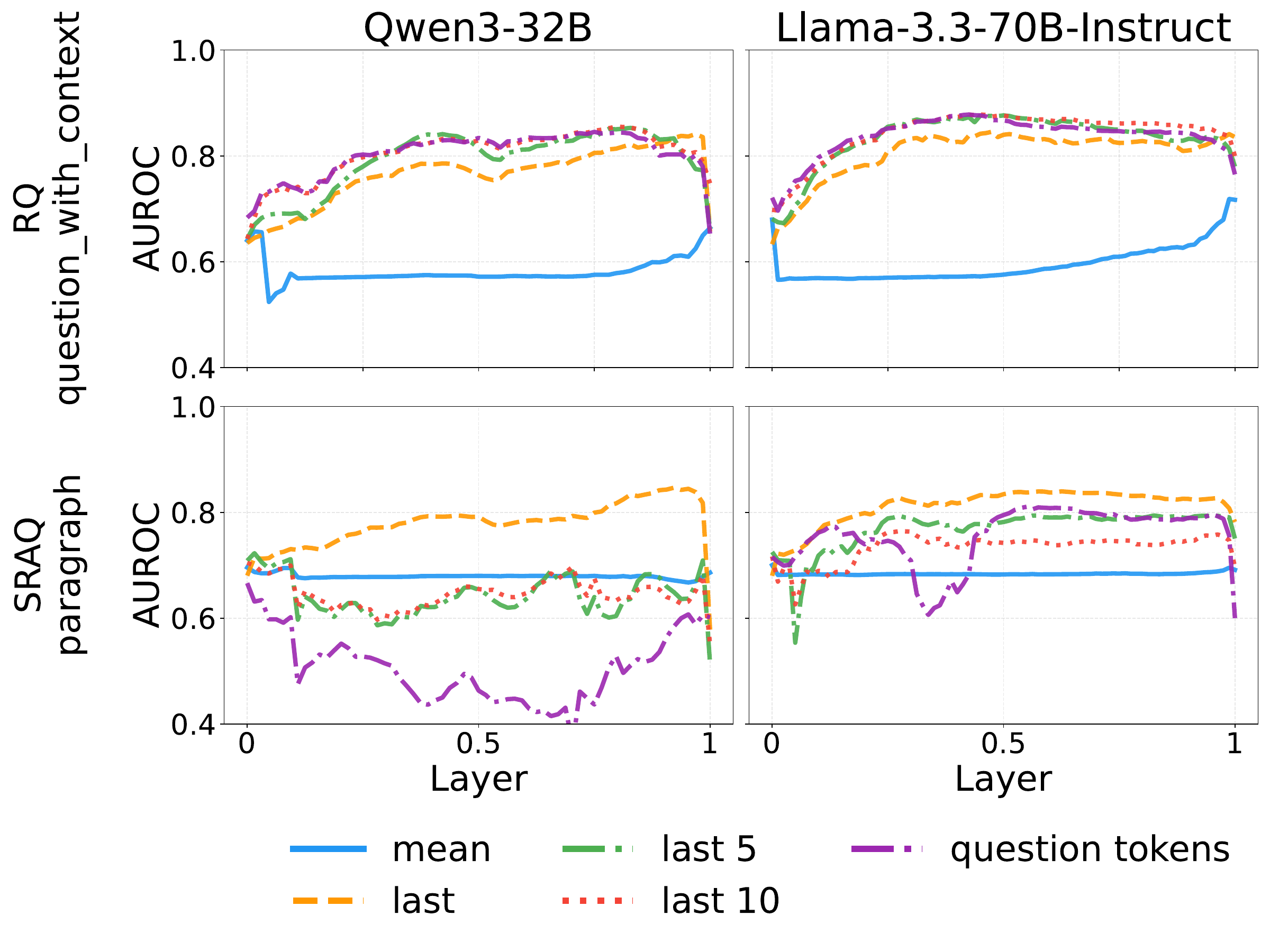}
    \caption{Test AUROC of training-free diffMean directions across layers under alternative pooling strategies. Results are shown for RQ (\emph{question\_with\_context}) and SRAQ (\emph{paragraph}) using Qwen3-32B and Llama-3.3-70B-Instruct. We compare mean pooling over all tokens, last-token pooling, pooling over the last 5 or 10 tokens, and mean pooling over question tokens only.}
    \label{fig:aggregation}
\end{figure}

To further examine the effect of representation choice, we run additional training-free diffMean experiments with alternative pooling strategies, shown in Figure~\ref{fig:aggregation}. In addition to mean pooling over all tokens and standard last-token pooling, we consider pooling over the last 5 or 10 tokens and mean pooling restricted to the question span only (\emph{question tokens}). We evaluate these variants across layers for Qwen3-32B and Llama-3.3-70B-Instruct on RQ (\emph{question\_with\_context}) and SRAQ (\emph{paragraph}). On RQ, pooling over the last few tokens and question-token pooling are often competitive with, and sometimes slightly better than, last-token pooling at middle layers. On SRAQ, these alternatives are generally less stable and often underperform last-token pooling, especially for Qwen3-32B. Overall, these results support the main-text focus on mean pooling and last-token representations as the two standard sequence-level settings for analyzing how rhetorical intent is encoded.

\section{Encoder-Based Model Results}
\label{app:bert}
\begin{figure}
    \centering
    \includegraphics[width=0.48\textwidth]{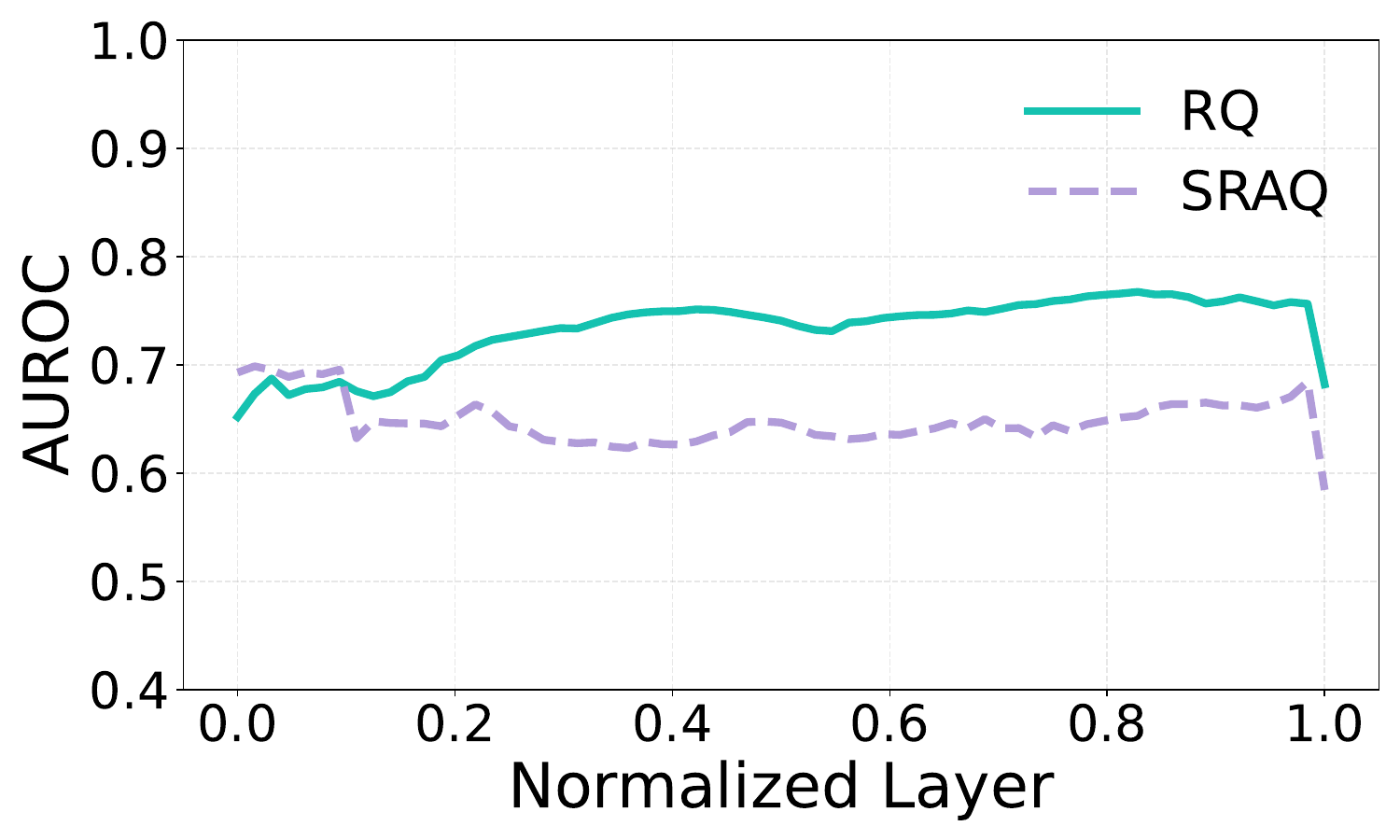}
    \caption{Test AUROC of training-free diffMean directions across normalized layers for ModernBERT-large using the [CLS] representation, shown for RQ and SRAQ.}
    \label{fig:bert}
\end{figure}

As a comparison to the decoder-only models in the main text, we also evaluate an encoder-based model, ModernBERT-large \citep{warner2025smarter}, using the [CLS] representation across layers. The results are shown in Figure~\ref{fig:bert}. On RQ, AUROC improves gradually in the earlier and middle layers, reaching a peak of 64.8 at layer 17, and then declines toward later layers. On SRAQ, performance is more stable across the network, remaining in a relatively narrow range between 67.8 and 69.9. Overall, these results remain below those of the decoder-only models reported in the main text. One possible interpretation is that rhetorical information is organized differently in encoder-based models, so that a single [CLS] representation may not provide the most effective readout for this task. More generally, the contrast suggests that representation choice may play a more important role for encoder-based models, where rhetorical cues could be distributed across tokens rather than concentrated in a single sequence-level summary. A broader encoder-decoder comparison, including alternative token aggregation strategies, is left for future work.

\section{Causal Steering}
\label{app:steering}

To validate that the identified directions capture rhetorical intent, we conduct causal steering experiments and report the results in this section, following the activation steering methodology of \citet{turner2024steering} and \citet{panickssery2024steering}.

\paragraph{Steering Mechanism.}
During inference, we apply steering by modifying the residual stream at a chosen layer. The steered hidden state is computed as
\begin{equation}
    h'_l = h_l + \alpha \cdot v_l ,
\label{eq:steering}
\end{equation}
where $h_l$ denotes the original residual stream at layer $l$, $h'_l$ the steered residual stream, $\alpha$ controls the strength of the intervention, and $v_l$ is the steering vector for that layer. In our experiments, we use the diffMean vectors identified in the main text.

\paragraph{Setup.}
We conduct our steering experiments on the RQ dataset using Qwen3-32B, where the context and the question are explicitly separated. Experiments are performed across seven layers, using 400 contexts drawn from the dataset (200 originally labeled as rhetorical and 200 as informational). To test whether the identified direction captures rhetorical question intent, we provide the model with only the context and prompt it to generate a follow-up question. The prompt template is shown in Figure~\ref{fig:steering-prompt}, and the generation hyperparameters are listed in Table~\ref{tab:steering-params}.

\begin{figure}[ht!]
\centering
\begin{tcolorbox}[
  colback=gray!5,
  colframe=gray!40,
  boxrule=0.5pt,
  arc=2pt,
  left=6pt,right=6pt,top=6pt,bottom=6pt
]
\footnotesize\ttfamily
\{context\}\\
Ask one concise follow-up question (ideally under 15 words). Your entire reply should be just that single question.
\end{tcolorbox}
\caption{Prompt template used for question generation in the steering experiments.}
\label{fig:steering-prompt}
\end{figure}

\begin{table}[t]
\centering
\small
\begin{tabular}{ll}
\toprule
\textbf{Parameter} & \textbf{Value} \\
\midrule
max\_new\_tokens      & 50   \\
do\_sample            & True \\
temperature           & 0.7  \\
top\_p                & 0.9  \\
repetition\_penalty   & 1.1  \\
\bottomrule
\end{tabular}
\caption{Generation parameters used in the steering experiments.}
\label{tab:steering-params}
\end{table}

\begin{figure}[ht]
    \centering
    \begin{subfigure}[t]{0.96\linewidth}
        \centering
        \includegraphics[width=\linewidth]{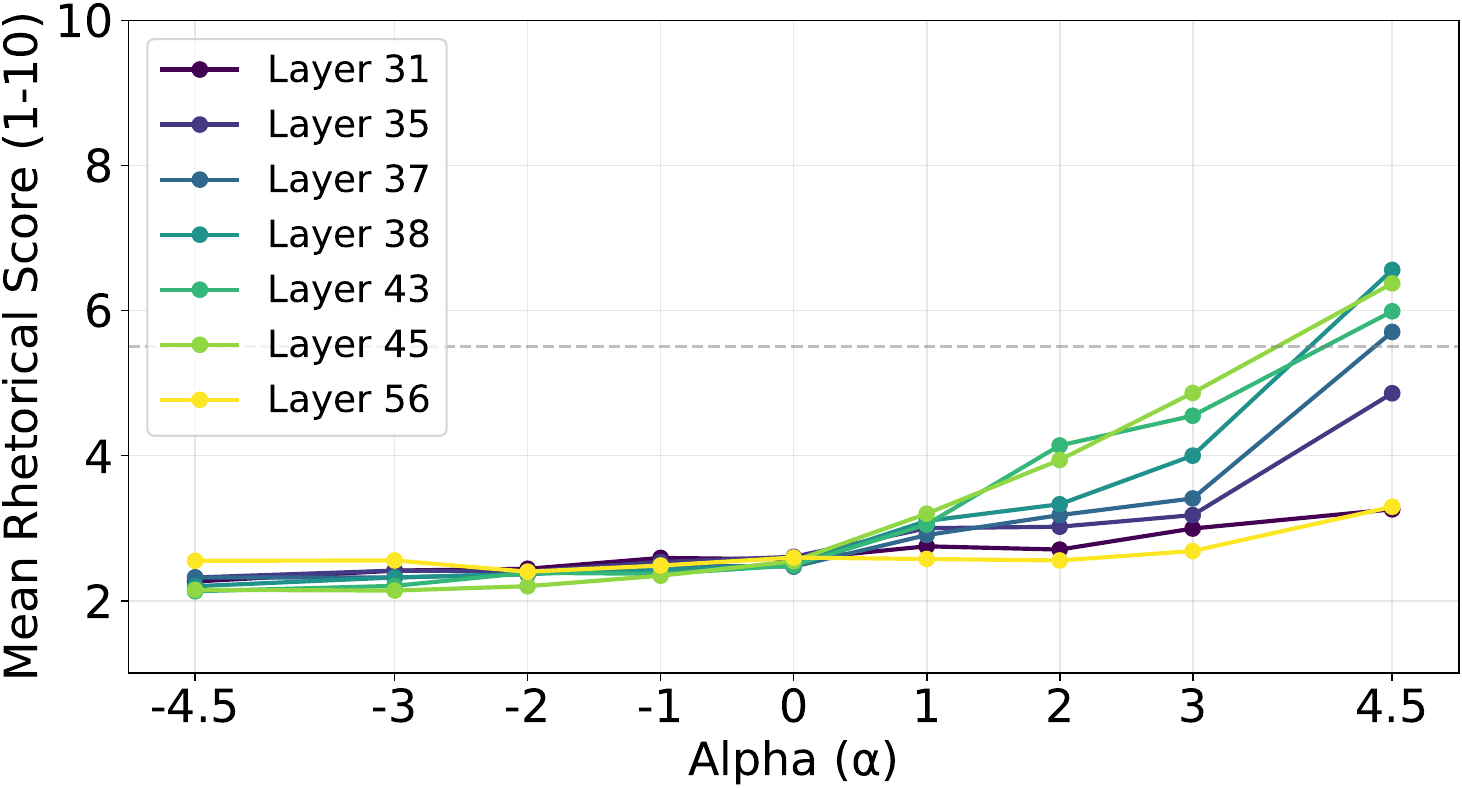}
        \caption{Contexts paired with rhetorical questions in the original dataset}
        \label{fig:alpha-sweep-rhetorical}
    \end{subfigure}

    \vspace{0.5em}

    \begin{subfigure}[t]{0.96\linewidth}
        \centering
        \includegraphics[width=\linewidth]{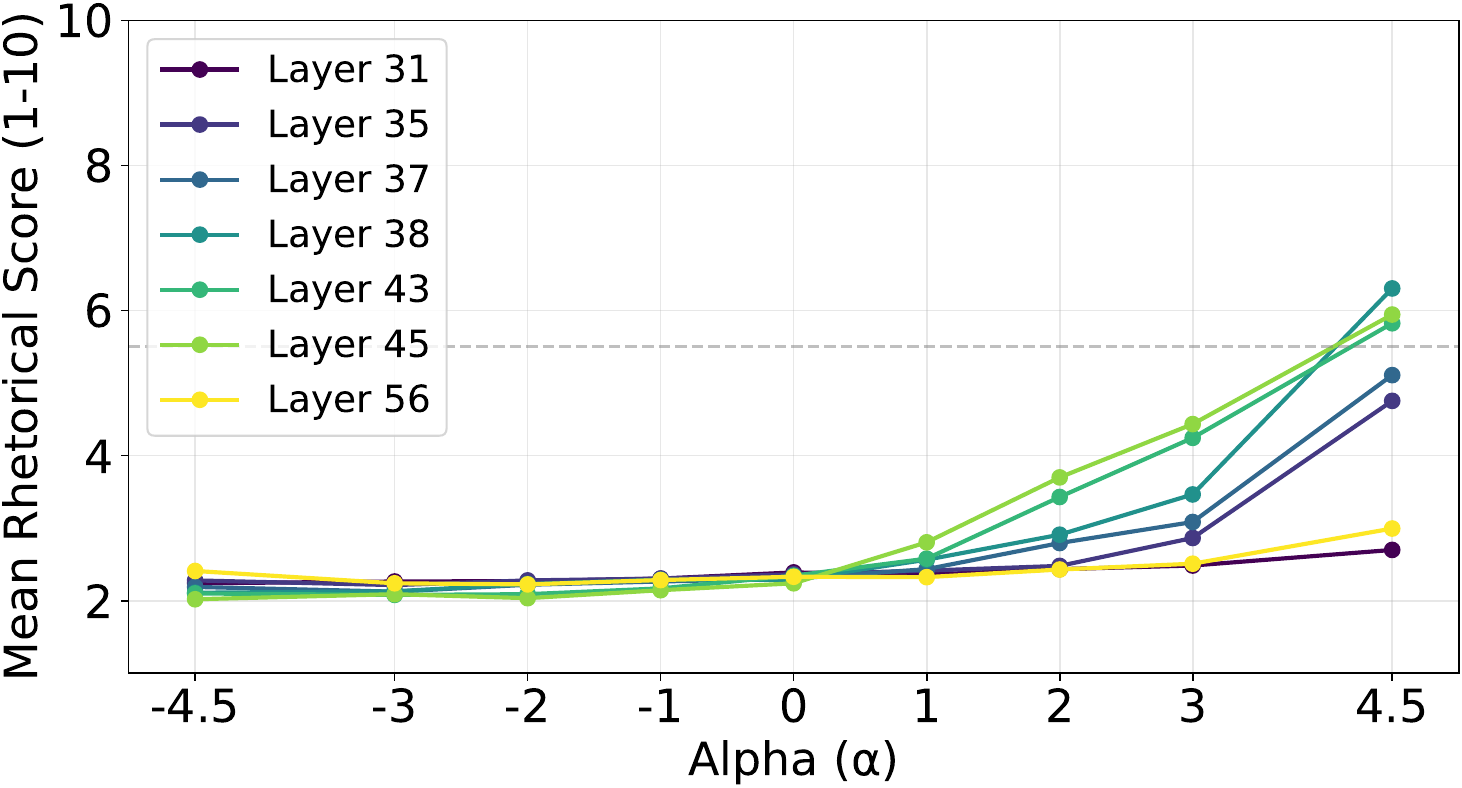}
        \caption{Contexts paired with informational questions in the original dataset}
        \label{fig:alpha-sweep-informational}
    \end{subfigure}

    \caption{Alpha sweep results on Qwen3-32B. Mean rhetorical score (y-axis) as a function of the steering coefficient $\alpha$ (x-axis) for different steering layers, evaluated separately on rhetorical and informational contexts.}
    \label{fig:alpha-sweep-combined}
\end{figure}

We then feed each generated question, together with its context, into GPT-5.1 to obtain a rhetorical score on a 1--10 scale, where 1 corresponds to a purely informational question and 10 corresponds to a strongly rhetorical question. The scoring prompt used for this evaluation is shown in Figure~\ref{fig:scoring-prompt}.

\begin{figure*}[ht!]
\centering
\begin{tcolorbox}[
  colback=gray!5,
  colframe=gray!40,
  boxrule=0.5pt,
  arc=2pt,
  left=6pt,right=6pt,top=6pt,bottom=6pt
]
\footnotesize\ttfamily
Your task is to rate a generated question on a scale of 1--10 based on how rhetorical versus informational it is.\\[0.5em]

\textbf{Scoring Scale:}\\
10 = Most rhetorical: strong presuppositions, obvious implied answers, highly persuasive tone.\\
7--9 = Rhetorical: contains presuppositions, implied answers, persuasive tone, or tag questions.\\
4--6 = Neutral / mixed: contains both factual and rhetorical elements.\\
2--3 = Informational: direct, factual question seeking specific information.\\
1 = Most informational: purely factual question with no rhetorical elements.\\[0.5em]

Context: \{context\}\\
Generated Question: \{question\}\\[0.5em]

Provide \textbf{only} a single integer from 1--10 as your rating.
\end{tcolorbox}
\caption{Prompt used to score generated questions on a rhetorical--informational scale.}
\label{fig:scoring-prompt}
\end{figure*}

\paragraph{Results.}
The steering results are shown in Figure~\ref{fig:alpha-sweep-combined}. Rhetorical scores increase largely monotonically with the steering strength $\alpha$, and for sufficiently large positive values (e.g., $\alpha = 4.5$), several layers reach scores around 7, indicating strongly rhetorical behavior. The magnitude of this effect varies across layers, with layers 38, 43, and 45 responding most strongly to steering, while others (e.g., layers 31 and 56) show weaker responses. Contexts originally paired with rhetorical questions consistently reach slightly higher scores, reinforcing the role of discourse context in shaping rhetorical interpretation. Taken together, these results provide evidence that the identified linear directions capture rhetorical question intent in the model’s internal representations.

\section{Additional Results on AUROC and Alignment}
\label{app:additional_results}

\begin{figure*}[ht!]
    \centering
    \begin{subfigure}[t]{\linewidth}
        \centering
        \includegraphics[width=\linewidth]{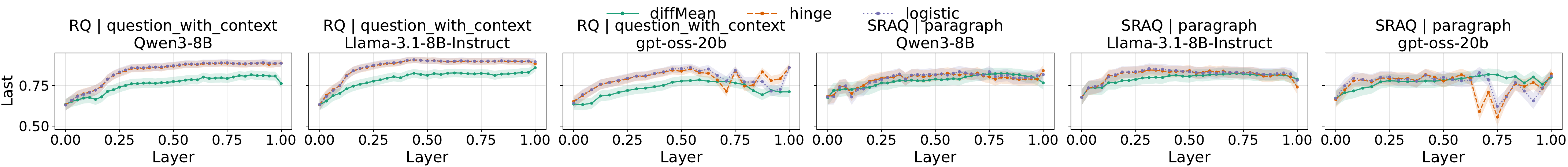}
        \caption{Additional LLMs. Last-token test AUROC across layers for models not shown in the main text.}
        \label{fig:appendix_lasttoken_more_models}
    \end{subfigure}
    \hfill
    \begin{subfigure}[t]{\linewidth}
        \centering
        \includegraphics[width=\linewidth]{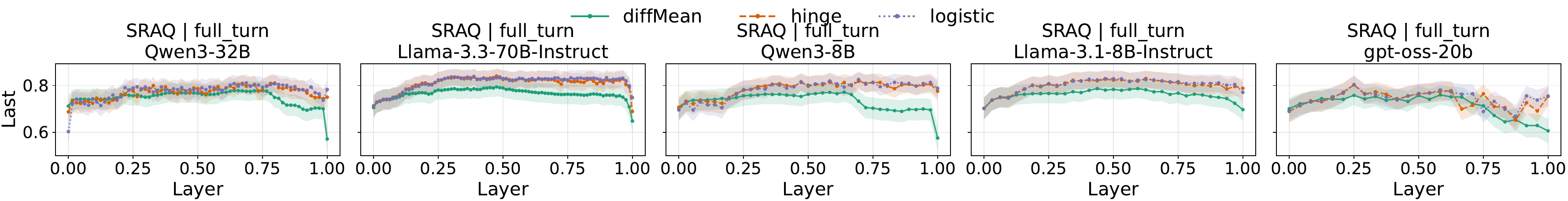}
        \caption{Additional input column. Last-token test AUROC across layers for the \texttt{full\_turn} formulation (not shown in the main text).}
        \label{fig:appendix_lasttoken_fullturn}
    \end{subfigure}

    \caption{Additional last-token probe results.
    Both panels report within-setting performance (training and evaluation within the same dataset, model, and input formulation) as a function of layer.
    (a) Results for additional LLMs beyond those highlighted in the main figures.
    (b) Results for the \texttt{full\_turn} input formulation.
    Across settings, trends are consistent with the main text: rhetorical question intent remains linearly separable across layers, and last-token representations provide stable signals.
    }
    \label{fig:appendix_lasttoken_additional}
\end{figure*}

\begin{figure*}[ht!]
    \centering
    \includegraphics[width=\linewidth]{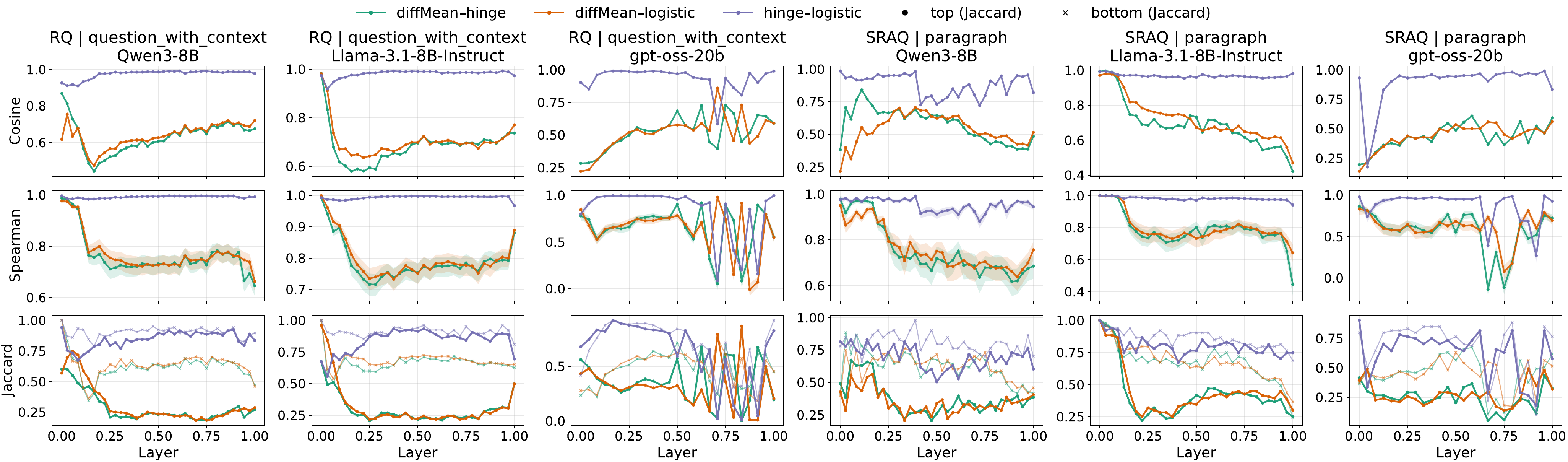}
    \caption{Within-dataset alignment across additional LLMs.
    Comparison of probe alignment metrics computed within the same dataset for models not shown in the main text.
    Although probes achieve similar discriminative performance, their learned directions exhibit limited alignment, indicating that multiple, non-collinear directions can support linear separability even within a fixed dataset.
    Trends are consistent with the main results.}
    \label{fig:app_within_dataset_alignment_other_models}
\end{figure*}

\begin{figure*}[t]
    \centering
    \includegraphics[width=\linewidth]{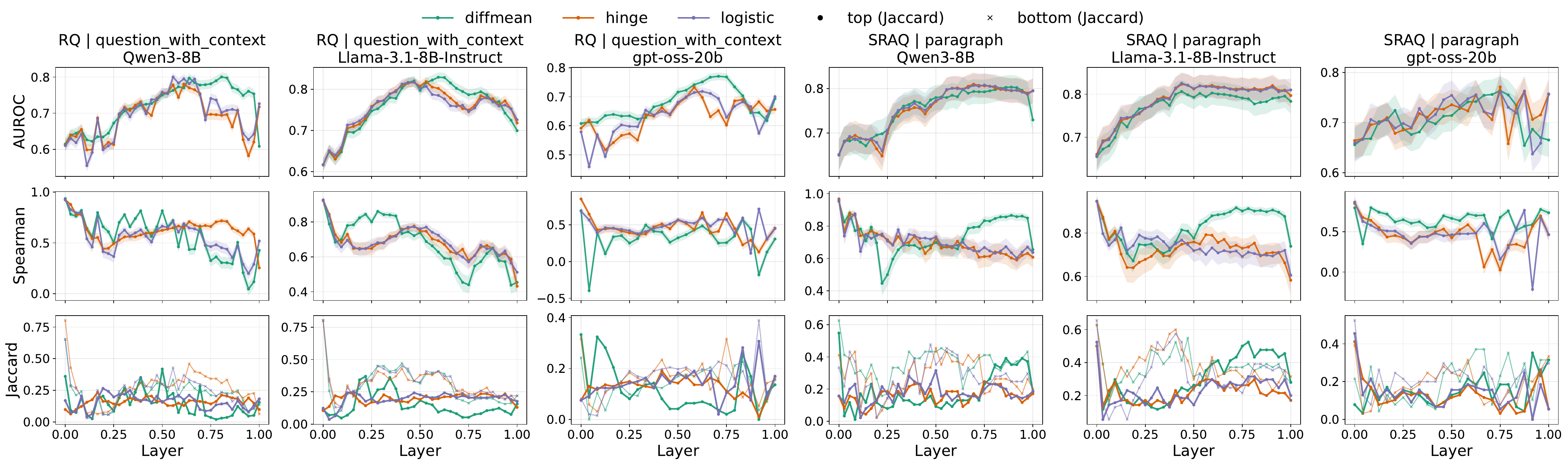}
    \caption{Cross-dataset alignment across additional LLMs.
    Comparison of probe alignment metrics when probe directions learned on one dataset are applied to another, for models not shown in the main text.
    Despite partial transferability in discriminative performance, alignment between probe directions remains limited, reflecting divergence in the induced rankings across datasets.
    These results are consistent with the trends reported in the main text.}
    \label{fig:app_transfer_alignment_other_models}
\end{figure*}

Figures~\ref{fig:appendix_lasttoken_additional}--\ref{fig:app_transfer_alignment_other_models} report additional analyses that extend the main text to settings not previously shown.
Specifically, we evaluate last-token probe performance on the \texttt{full\_turn} input formulation for SRAQ, and examine both within-dataset and cross-dataset alignment for three additional LLMs: \textsc{Qwen3-8B}, \textsc{Llama-3.1-8B-Instruct}, and \textsc{GPT-OSS-20B}.

Across these settings, rhetorical questions remain linearly separable, with AUROC trends broadly consistent with those reported in the main text.
However, results for \textsc{GPT-OSS-20B} exhibit noticeably higher variability across layers and runs compared to the other models, both in AUROC and in alignment-based measures.
We do not observe comparable instability for \textsc{Qwen3-8B} or \textsc{Llama-3.1-8B-Instruct}.
While we do not investigate the source of this behavior further, this observation suggests that probe stability may vary substantially across model architectures, even when overall discriminative performance appears similar.

\end{document}